\documentclass{article}

\usepackage[main, final]{neurips_2026}

\usepackage[utf8]{inputenc} 
\usepackage[T1]{fontenc}    
\usepackage{fix-cm}         
\usepackage{hyperref}       
\usepackage{url}            
\usepackage{booktabs}       
\usepackage{amsmath}        
\usepackage{amsfonts}       
\usepackage{nicefrac}       
\usepackage{microtype}      
\usepackage{xcolor}         
\usepackage{enumitem}       
\usepackage{tikz}           
\usetikzlibrary{arrows.meta,positioning,calc}
\usepackage{pgfplots}       
\pgfplotsset{compat=1.18}
\usepgfplotslibrary{groupplots}
\usepackage{algorithm}      
\usepackage{algorithmic}    
\setcitestyle{round}        

\title{AMARIS: Merging Generalist and Specialist LLMs via Adaptive Subspace Inheritance}

\author{%
  Hwee Young Kang \\
  DSO National Laboratories\\
  Singapore \\
  \texttt{khweeyou@dso.org.sg} \\
  \And
  Irwin De Tao Chin \\
  DSO National Laboratories\\
  Singapore \\
  \texttt{cdetao@dso.org.sg} \\
  \And
  Wen-Haw Chong \\
  DSO National Laboratories\\
  Singapore \\
  \texttt{cwenhaw@dso.org.sg} \\
  \And
  Hai Leong Chieu \\
  DSO National Laboratories\\
  Singapore \\
  \texttt{chaileon@dso.org.sg} \\
}

\begin{document}

\maketitle

\begin{abstract}
Domain specialization of large language models (LLMs) often improves target-domain performance but degrades broad capabilities. Model merging is an attractive training-free alternative, yet common parameter-space strategies (for example, interpolation and sparsified averaging) provide limited control over task-specific interference. We introduce Adaptive Model Alignment via Ranked Inherited Subspaces (AMARIS), a covariance-driven framework that transfers specialist updates only along selected activation directions. AMARIS formulates merging as budgeted subspace inheritance: it seeks high specialist activation capture under a regularized base-output perturbation budget, then builds a low-rank projection gate from ranked generalized eigendirections. This yields a single static merged checkpoint that concentrates transfer in high-utility, low-interference directions. We evaluate AMARIS on Qwen3-based merges (8B and 32B) across finance, instruction-following, and multilingual SEA specialization. In our evaluated settings, AMARIS reaches Pareto operating points that are competitive with strong training-free baselines, including 98.37\% specialist retention with 95.16\% core retention for the 8B finance setting, 103.55\% core retention with 90.28\% specialist retention for the 8B instruction-following setting, and 99.20\% core retention with 94.80\% specialist retention for the 32B SEA setting. These results support subspace-constrained merging as a practical route to compositional capability integration without additional training.
\end{abstract}

\noindent Code is available at \url{https://github.com/adriankhy93/AMARIS-neurips}.

\section{Introduction}
Transformer-based large language models (LLMs) \citep{transformer} are increasingly deployed as general-purpose assistants, yet production use often requires a difficult combination: broad general capability together with high-quality domain specialization. A standard adaptation strategy is continued pre-training or fine-tuning from a strong base checkpoint \citep{continual,lora}. In practice, however, sequential specialization often induces catastrophic forgetting \citep{ewc,forgetting}: target-domain performance improves while previously strong capabilities regress. This creates a persistent systems tension between specialization and retention.

Model merging is an attractive training-free alternative. Given a base model and specialized descendants, the objective is a single checkpoint that inherits specialist gains without sacrificing base competence. Existing approaches such as model soups, SLERP, TIES, and DARE-TIES \citep{soup,slerp,ties,dareties} can be effective, but they primarily manipulate parameters directly and do not provide data-conditioned control over \emph{which} directions should transfer. When source models are tuned to distinct distributions, this can yield brittle specialist--generalist trade-offs.

We propose \textbf{AMARIS} (\textbf{A}daptive \textbf{M}odel \textbf{A}lignment via \textbf{R}anked \textbf{I}nherited \textbf{S}ubspaces), a covariance-driven merging framework that conditions transfer on activation geometry. At each layer, AMARIS estimates base and specialist activation covariances from small calibration sets, then selects inherited directions that favor high specialist activation capture under an explicit budget on regularized base-output perturbation. This yields a budgeted subspace selection problem approximated using generalized eigendirection ranking with greedy budgeted acceptance. The resulting low-rank projector gates the parameter delta so transfer concentrates in high-utility, low-interference directions.

\begin{figure}[t]
\centering
\begingroup
\definecolor{figink}{HTML}{182B42}
\definecolor{figblue}{HTML}{2166AC}
\definecolor{figorange}{HTML}{AD5315}
\definecolor{figgreen}{HTML}{167467}
\definecolor{figgray}{HTML}{65758A}
\begin{tikzpicture}[x=0.33pt,y=-0.33pt,inner sep=0pt,outer sep=0pt]
\path[use as bounding box] (0,0) rectangle (1200,245);
\node[anchor=base west,text=figink,font={\sffamily\fontsize{9.57}{11.48}\selectfont\bfseries}] at (10,30) {Layerwise merge};
\path[draw=figblue,fill=figblue!5,line width=0.65pt,rounded corners=2.6pt] (10,60) rectangle (290,235);
\node[anchor=base west,text=figink,font={\sffamily\fontsize{9.57}{11.48}\selectfont\bfseries}] at (26,96) {Calibration};
\path[draw=figblue,fill=figblue!5,line width=0.65pt,rounded corners=2.6pt] (335,60) rectangle (615,235);
\node[anchor=base west,text=figink,font={\sffamily\fontsize{8.58}{10.30}\selectfont\bfseries}] at (351,96) {Select subspace};
\path[draw=figgreen,fill=figgreen!5,line width=0.65pt,rounded corners=2.6pt] (660,60) rectangle (950,235);
\node[anchor=base west,text=figink,font={\sffamily\fontsize{9.57}{11.48}\selectfont\bfseries}] at (676,96) {Merge weights};
\path[draw=figgreen,fill=figgreen!5,line width=0.65pt,rounded corners=2.6pt] (995,60) rectangle (1190,235);
\node[anchor=base west,text=figink,font={\sffamily\fontsize{8.58}{10.30}\selectfont\bfseries}] at (1011,96) {Checkpoint};
\node[anchor=base west,text=figblue,font={\sffamily\fontsize{8.91}{10.69}\selectfont}] at (26,151) {$W_1\to X_1\to C_1$};
\node[anchor=base west,text=figorange,font={\sffamily\fontsize{8.91}{10.69}\selectfont}] at (26,199) {$W_2\to X_2\to C_2$};
\node[anchor=base west,text=figink,font={\sffamily\fontsize{8.58}{10.30}\selectfont}] at (351,144) {Rank + budget $\alpha$};
\node[anchor=base west,text=figink,font={\sffamily\fontsize{8.58}{10.30}\selectfont}] at (351,181) {Select subspace $U$};
\node[anchor=base west,text=figink,font={\sffamily\fontsize{9.24}{11.09}\selectfont}] at (351,217) {$P=UU^\top$};
\node[anchor=base west,text=figink,font={\sffamily\fontsize{8.91}{10.69}\selectfont}] at (676,166) {$W_1+(W_2-W_1)P$};
\node[anchor=base west,text=figink,font={\sffamily\fontsize{8.58}{10.30}\selectfont}] at (1011,166) {Static model};
\draw[draw=figgray,line width=0.8pt,-{Latex[length=4pt,width=3pt]}] (290,147.5) -- (330,147.5);
\draw[draw=figgray,line width=0.8pt,-{Latex[length=4pt,width=3pt]}] (615,147.5) -- (655,147.5);
\draw[draw=figgray,line width=0.8pt,-{Latex[length=4pt,width=3pt]}] (950,147.5) -- (990,147.5);
\end{tikzpicture}
\endgroup
\caption{\textbf{AMARIS overview.} Each checkpoint supplies layer-input statistics on its calibration set. AMARIS ranks candidate directions, selects a subspace under a preservation budget, and folds the projected delta into a static checkpoint.}
\label{fig:amaris_overview}
\label{fig:projection_mechanism}
\end{figure}
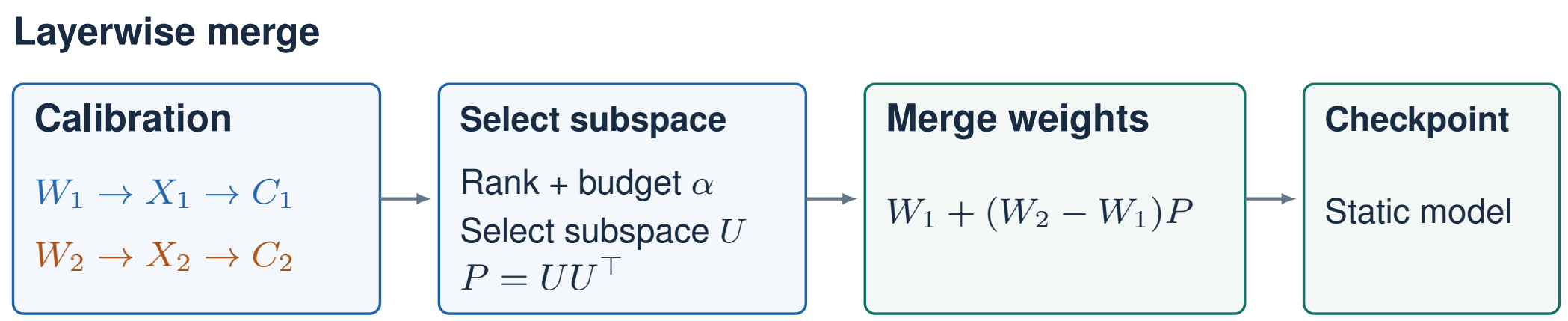

Figure~\ref{fig:amaris_overview} summarizes the pipeline.

AMARIS replaces a single global interpolation coefficient with layerwise geometric decisions: directions with high specialist utility and low base interference are kept, while high-interference directions are filtered out. Relative to prior training-free heuristics, its distinguishing feature is explicit optimization of a data-conditioned objective with a user-controlled preservation budget.

\paragraph{Contributions.}
\begin{itemize}
  \item \textbf{Objective-level formulation.} We cast checkpoint merging as budgeted subspace inheritance, where specialist capture is maximized under an explicit base-overlap constraint defined on activation covariances.
  \item \textbf{Training-free algorithm.} We introduce AMARIS, a covariance-driven layerwise merge procedure that ranks generalized eigendirections and applies projection-gated delta transfer under a preservation budget.
  \item \textbf{Empirical trade-off quality.} Across Qwen3-8B and Qwen3-32B \citep{qwen3} merges, AMARIS reaches competitive core--specialist operating points. In the complete SEAHELM evaluation, it improves Tamil by 4.34 points over the base while remaining below the specialist, so this is partial rather than full transfer. We also show that the gain is statistically significant.
  \item \textbf{Practical retention profile.} In the three evaluated settings, AMARIS preserves at least 90\% specialist capability and at least 95\% base core capability.
\end{itemize}

Together, these results position subspace-constrained merging as a principled and scalable path for compositional capability integration in LLM systems.
\section{Related Work and Positioning}
\label{sec:related}

\paragraph{Continual adaptation and forgetting.}
A standard route to domain specialization is to continue training from a strong general-purpose checkpoint \citep{continual}. Although this often improves target-domain behavior, it can also induce capability regression when optimization shifts model capacity toward a new distribution \citep{forgetting}. This pattern is consistent with continual-learning evidence on catastrophic forgetting and cross-task interference \citep{ewc}. In our setting, these observations motivate training-free recombination methods that recover specialist gains while preserving base capabilities.

\paragraph{LLM model merging.}
Checkpoint merging seeks one deployable model from multiple checkpoints with shared architecture and parameterization. Existing methods include interpolation (model soups \citep{soup}, SLERP \citep{slerp}), sparsification/sign heuristics (TIES and DARE-TIES \citep{ties,dareties}), and data-aware fusion variants such as RegMean \citep{jin2023dataless} and LEWIS-guided sparse merging \citep{lewis}. These methods are practical and training-free, but they offer limited explicit, data-conditioned control over capability-specific transfer at merge time. AMARIS is complementary: it selects transfer directions from calibration-derived activation geometry under an explicit preservation budget.

\paragraph{Subspace and activation perspectives.}
Adaptation updates are often low-dimensional \citep{lora}, and representation interventions can induce targeted behavioral changes \citep{steering,rome}. AMARIS follows this representation-centric view but remains deployment-simple: no trainable adapters, no learned routers, and no inference-time conditionals, only static low-rank projection gates estimated from calibration activations.

\paragraph{Positioning AMARIS.}
Across these lines of work, two objectives are often in tension: practical training-free merging and capability-specific control. AMARIS targets this intersection by formulating merge-time transfer as budgeted subspace inheritance over activation covariances. This yields an explicit objective for specialist capture versus base overlap, a transparent control knob ($\alpha$), and a single static merged checkpoint with no additional training stage.

\section{Problem Formulation}
\label{sec:setup}

We study \emph{capability recombination} between two checkpoints with identical architecture and tokenization: a base model $M_1$ to preserve on general tasks, and a specialized model $M_2$ to inherit on a target domain. The objective is a single merged model $M_{\text{merge}}$ that improves target-domain behavior relative to $M_1$ while limiting regression on base capabilities.

For a batch of layer inputs $X\in\mathbb{R}^{N\times d}$ and linear weights $W\in\mathbb{R}^{d_{\mathrm{out}}\times d}$, we use
\[
Y = XW^\top.
\]
All merge decisions are therefore expressed through their effect on outputs $Y$.

We first consider one linear layer with weights $W_1$ and $W_2$ from the base and specialist models, respectively:
\[
W_1, W_2 \in \mathbb{R}^{d_{\mathrm{out}}\times d}, \qquad \Delta = W_2 - W_1.
\]
Rather than fully replacing $W_1$ with $W_2$, we seek a subspace-restricted transfer
\[
W_{\text{merge}} = W_1 + \Delta P,
\]
where $P \in \mathbb{R}^{d\times d}$ is an orthogonal projector ($P=P^\top=P^2$). To see why this preserves base behavior outside the selected subspace, write
\[
Y_{\text{merge}}=XW_{\text{merge}}^\top = X(W_1+\Delta P)^\top = XW_1^\top + XP\Delta^\top.
\]
Only the projected input component $XP$ is affected by the specialist delta $\Delta$. In particular, if a row activation $x$ satisfies $xP=0$, then
\[
xW_{\text{merge}}^\top = xW_1^\top,
\]
so that component remains exactly anchored to $M_1$.

To determine which directions should transfer, we collect layer inputs $X_1\in\mathbb{R}^{N_1\times d}$ from a base-preferred calibration set with model $M_1$ and $X_2\in\mathbb{R}^{N_2\times d}$ from a specialist-preferred calibration set with model $M_2$, respectively. Define
\[
C_1=\frac{1}{N_1}X_1^\top X_1,\qquad
C_2=\frac{1}{N_2}X_2^\top X_2,\qquad
G=\Delta^\top\Delta.
\]
At fixed calibration inputs, base preservation and specialist transfer are measured by the mean squared layer-output errors
\begin{align}
L_{\mathrm{base}}(P)
&=\frac{1}{N_1}\|Y_{\mathrm{merge}}(X_1)-Y_1(X_1)\|_F^2
=\frac{1}{N_1}\|X_1P\Delta^\top\|_F^2
=\mathrm{tr}(PC_1PG),
\label{eq:base_preservation}\\
L_{\mathrm{specialist}}(P)
&=\frac{1}{N_2}\|Y_{\mathrm{merge}}(X_2)-Y_2(X_2)\|_F^2
\nonumber\\
&=\frac{1}{N_2}\|X_2(I-P)\Delta^\top\|_F^2
=\mathrm{tr}((I-P)C_2(I-P)G).
\label{eq:specialist_transfer}
\end{align}
We express their trade-off as
\begin{equation}
\min_{P=P^\top=P^2}\;L_{\mathrm{specialist}}(P)
\quad\text{s.t.}\quad
L_{\mathrm{base}}(P)\le\alpha\,\mathrm{tr}(C_1G),
\qquad \alpha\in[0,1].
\label{eq:exact_objective}
\end{equation}
Here $\mathrm{tr}(C_1G)$ is the base perturbation under full transfer ($P=I$), so $\alpha$ sets the allowed fraction of that perturbation. This formulation also covers a zero full-transfer reference without dividing by it. The constraint concerns a single layer at fixed inputs, rather than an end-to-end capability guarantee.

The exact errors couple the selected directions through $G$. To construct inexpensive candidate rankings, we instead use activation-energy bounds. By $\|AB\|_F\le\|A\|_F\|B\|_2$,
\begin{align*}
L_{\mathrm{base}}(P)
&\le\|\Delta\|_2^2\,\mathrm{tr}(PC_1),\\
L_{\mathrm{specialist}}(P)
&\le\|\Delta\|_2^2\bigl[\mathrm{tr}(C_2)-\mathrm{tr}(PC_2)\bigr].
\end{align*}
Since $\mathrm{tr}(C_2)$ is constant with respect to $P$, maximizing $\mathrm{tr}(PC_2)$ minimizes the specialist bound. These bounds motivate the activation-energy surrogate
\begin{equation}
\max_{P=P^\top=P^2}\;\mathrm{tr}(PC_2)
\quad\text{s.t.}\quad
\mathrm{tr}(PC_1)\le\alpha\,\mathrm{tr}(C_1).
\label{eq:constrained_objective}
\end{equation}
The normalized cost is the fraction of base activation energy captured by the subspace. This surrogate motivates ranking, but its normalized constraint is not equivalent to the output-perturbation constraint in \eqref{eq:exact_objective}. AMARIS checks the delta-dependent cost during candidate acceptance, as described next.

Appendix~\ref{app:setup_derivation} provides the full derivation of the output-error expressions and activation-energy bounds.

\section{Method: AMARIS}
\label{sec:method}

AMARIS constructs a transfer subspace independently for each targeted linear weight using calibration statistics, spectral ranking, and budgeted greedy selection. Ranking uses the activation-energy surrogate from Equation~\ref{eq:constrained_objective}; acceptance uses the delta-dependent preservation cost from Equation~\ref{eq:exact_objective}.

\paragraph{From objective to direction scoring.}
Write $P=UU^\top$ with $U^\top U=I$. Then
\[
\mathrm{tr}(PC_2)=\mathrm{tr}(U^\top C_2U),\qquad
\mathrm{tr}(PC_1)=\mathrm{tr}(U^\top C_1U).
\]
For a single unit direction $u$, these quantities are specialist benefit $u^\top C_2u$ and base activation cost $u^\top C_1u$. Their ratio prioritizes specialist energy relative to base energy. With ridge stabilization, the scale-invariant ranking score is
\[
\rho(u)=\frac{u^\top C_2u}{u^\top(C_1+\eta I)u},\qquad\eta>0.
\]
Its stationary directions satisfy the generalized eigenproblem
\[
C_2u_i=\rho_i(C_1+\eta I)u_i.
\]
We approximate the leading eigenpairs and retain up to $m$ candidates in decreasing estimated eigenvalue order. This score supplies a candidate-ranking heuristic; it does not solve either constrained projector problem globally.

\paragraph{Budgeted greedy subspace construction.}
We scan the retained candidate directions in decreasing estimated eigenvalue order and accept each direction only if the enlarged subspace satisfies the preservation budget. Here $m$ is the maximum number of candidates considered per layer, not the final subspace rank. Generalized eigenvectors are orthogonal in the $(C_1+\eta I)$-metric, but need not be Euclidean-orthogonal; we therefore orthonormalize the accepted directions in the Euclidean metric to construct $P=UU^\top$.

Let $U$ denote the current accepted basis, initially empty, and set $\tau=\alpha\,\mathrm{tr}(C_1G)$. For each candidate $u_i$, where $i$ indexes the ranked candidates:
\begin{enumerate}[leftmargin=*,itemsep=2pt,topsep=3pt]
\item Form the tentative Euclidean-orthonormal basis $U_{\mathrm{try}}=\mathrm{orth}([U,u_i])$, where $[U,u_i]$ appends the candidate to the current basis.
\item Construct $P_{\mathrm{try}}=U_{\mathrm{try}}U_{\mathrm{try}}^\top$ and evaluate the complete preservation cost:
\[
\mathrm{tr}(P_{\mathrm{try}}C_1P_{\mathrm{try}}G)\le\tau.
\]
\item If the inequality holds, accept the candidate by setting $U\leftarrow U_{\mathrm{try}}$. Otherwise retain the current $U$. In either case, continue to the next candidate.
\end{enumerate}
For numerical stability, the implementation replaces $C_1$ with $C_1+\eta I$ in both the generalized eigenproblem and the preservation-budget calculation; implementation details and the distinction from the unregularized constraint are given in Appendix~\ref{app:setup_derivation}.

Each candidate is tested using the cost of the entire enlarged subspace, accounting for interactions with previously accepted directions. After all retained candidates have been considered, the final projector is
\[
P_{\mathrm{AMARIS}}=U_{\mathrm{AMARIS}}U_{\mathrm{AMARIS}}^\top,
\]
where $U_{\mathrm{AMARIS}}$ is the final accepted basis. Its rank is determined by budgeted acceptance within the candidate pool; an empty basis gives $P=0$. Greedy selection enforces budget feasibility but carries no global-optimality or approximation guarantee.

\paragraph{Layerwise merge rule.}
For each projected weight, AMARIS applies
\[
W_{\mathrm{merge}}=W_1+(W_2-W_1)P_{\mathrm{AMARIS}}.
\]
Specialist updates are inherited only along selected directions, while complementary input directions retain the base weight contribution. The budget $\alpha$ controls which candidate directions can be accepted.

\paragraph{Remaining parameters.}
Projection applies to linear layer weights. All remaining parameters, including embeddings, normalization parameters, and biases, use
\[
\theta_{\mathrm{merge}}=\beta\theta_{\mathrm{base}}+(1-\beta)\theta_{\mathrm{specialist}}.
\]
Thus $\beta=1$ retains their base values, independently of $\alpha$. AMARIS requires covariance estimation, spectral solving, and budgeted selection, with no additional training, adapters, or inference-time routing.

\paragraph{Main algorithm.} Algorithm~\ref{alg:amaris} summarizes the merge procedure, with $\mathcal{L}$ denoting the targeted linear layers and $r_\ell\le m$ the number of retained candidates for layer $\ell$.

\begin{algorithm}[t]
\caption{\textsc{AMARIS}: merge procedure}
\label{alg:amaris}
\small
\begin{algorithmic}[1]
\REQUIRE base model $M_1$, specialist model $M_2$; calibration activations
$\{X_1^{(\ell)}, X_2^{(\ell)}\}_{\ell\in\mathcal{L}}$; budget $\alpha$; remaining-parameter weight $\beta$; ridge $\eta$; candidate cap $m$
\ENSURE merged model $M_{\mathrm{merge}}$

\STATE \textbf{Initialize.} $M_{\mathrm{merge}} \leftarrow M_1$

\STATE \textbf{(1) Compute layer covariances and deltas.}
\FOR{$\ell \in \mathcal{L}$}
  \STATE $\Delta^{(\ell)} \leftarrow W_2^{(\ell)} - W_1^{(\ell)}$
  \STATE $C_1^{(\ell)} \leftarrow \frac{1}{N_1}(X_1^{(\ell)})^\top X_1^{(\ell)}$
  \STATE $C_2^{(\ell)} \leftarrow \frac{1}{N_2}(X_2^{(\ell)})^\top X_2^{(\ell)}$
\ENDFOR

\STATE \textbf{(2) Rank candidate directions and set budgets.}
\FOR{$\ell \in \mathcal{L}$}
  \STATE Approximate leading eigenpairs of $C_2^{(\ell)}u_i = \rho_i(C_1^{(\ell)}+\eta I)u_i$; retain $r_\ell\le m$ candidates sorted by descending $\rho_i$, where $\eta$ is a small ridge parameter
  \STATE Store the ordered candidate list $\mathcal{V}^{(\ell)} \leftarrow (u_1,\ldots,u_{r_\ell})$
  \STATE $G^{(\ell)} \leftarrow (\Delta^{(\ell)})^\top\Delta^{(\ell)}$
  \STATE $B^{(\ell)} \leftarrow C_1^{(\ell)}+\eta I, \; \tau^{(\ell)} \leftarrow \alpha\,\mathrm{tr}(B^{(\ell)}G^{(\ell)})$, \; $U^{(\ell)} \leftarrow [\,]$
\ENDFOR

\STATE \textbf{(3) Budgeted subspace selection.}
\FOR{$\ell \in \mathcal{L}$}
  \FOR{each $u_i$ in the ordered candidate list $\mathcal{V}^{(\ell)}$}
    \STATE Orthogonalize $u_i$ against $U^{(\ell)}$; skip a numerically zero residual
    \STATE $U_{\mathrm{try}} \leftarrow \mathrm{orth}([U^{(\ell)},u_i])$, \; $P_{\mathrm{try}} \leftarrow U_{\mathrm{try}}U_{\mathrm{try}}^\top$
    \IF{$\mathrm{tr}(P_{\mathrm{try}} B^{(\ell)} P_{\mathrm{try}} G^{(\ell)}) \le \tau^{(\ell)}$}
      \STATE $U^{(\ell)} \leftarrow U_{\mathrm{try}}$
    \ENDIF
  \ENDFOR
\ENDFOR

\STATE \textbf{(4) Apply merge rule.}
\FOR{$\ell \in \mathcal{L}$}
  \STATE $U_{\mathrm{AMARIS}}^{(\ell)} \leftarrow U^{(\ell)}$
  \STATE $P_{\mathrm{AMARIS}}^{(\ell)} \leftarrow U_{\mathrm{AMARIS}}^{(\ell)}(U_{\mathrm{AMARIS}}^{(\ell)})^\top$
  \STATE $W_{\mathrm{merge}}^{(\ell)} \leftarrow W_1^{(\ell)} + \Delta^{(\ell)} P_{\mathrm{AMARIS}}^{(\ell)}$
\ENDFOR

\STATE Mix all remaining parameters as $\theta_{\mathrm{merge}}\leftarrow\beta\theta_{\mathrm{base}}+(1-\beta)\theta_{\mathrm{specialist}}$
\STATE \textbf{return} $M_{\mathrm{merge}}$
\end{algorithmic}
\end{algorithm}

\section{Experiments and Results}

\label{sec:experiments}

\subsection{Evaluation Protocol}

\paragraph{Key findings.}
AMARIS consistently occupies favorable core-specialist operating points in our evaluated settings. In finance, instruction-following, and SEA settings, it achieves the highest preservation of core capabilities while maintaining strong specialist performance. In the SEA setting, it also achieves the highest SEAHELM specialist retention (94.80\% normalized) among the compared merge baselines, with the largest improvement over the base occurring in Tamil.

We evaluate AMARIS in three two-model merge settings to test whether specialist capabilities can be transferred without sacrificing base-model competence. In all settings, the base checkpoint is a Qwen model and the specialist checkpoint is a domain-adapted fine-tuned derivative (or a regional specialist model).

\paragraph{Research questions.}
Our experiments are organized around three questions: 
\begin{enumerate}
  \item \textbf{RQ1} Can AMARIS improve the core--specialist frontier over strong training-free merge baselines?
  \item \textbf{RQ2} Is the preservation budget $\alpha$ a predictable control knob?
  \item \textbf{RQ3} Do improvements hold across heterogeneous specialization regimes (finance, instruction-following, multilingual SEA)?
\end{enumerate}

\paragraph{Evaluation metrics and reporting.}
Core capability retention is measured on AIME26 (math reasoning) \citep{aime2026}, LCB\_v6 (code reasoning) \citep{lcb}, IFEval (general instruction following) \citep{ifeval}, and GPQA:Diamond (knowledge reasoning) \citep{gpqa}. Specialist transfer is measured by FPB \citep{finance} in finance and IFBench \citep{ifbench} in instruction following; for the SEA setting, specialist retention is summarized by average SEAHELM \citep{seahelm} scores across languages. We use SEAHELM v1 with evaluation-mode extensions and a \texttt{gpt-oss-120b} \citep{gptoss} multi-turn judge, as detailed in Appendix~\ref{app:seahelm_implementation}.

\paragraph{Table conventions.}
In Tables~\ref{tab:set1_finance}--\ref{tab:set3_sea_core}, bracketed percentages normalize core scores to the base and specialist scores to the specialist checkpoint. Core Avg. is the arithmetic mean of the four core scores. Bold and underlining identify the best and second-best merging methods, excluding the source checkpoints; Core Avg. rankings use this ratio, with ties at the displayed precision receiving the same highlighting. Higher scores are better. All rankings are point estimates, with paired-bootstrap comparisons reported in Appendix~\ref{app:bootstrap}. The metrics reported for evaluations are stated in Appendix~\ref{app:reproducibility}.

\paragraph{Uncertainty reporting.}
We report pass@1:16 for AIME26 and LCB\_v6, and pass@1:8 for GPQA:Diamond to reduce variance. We report paired evaluation-set bootstrap confidence intervals for fixed checkpoints, using 10,000 question-level paired resamples and percentile 95\% intervals. Appendix~\ref{app:bootstrap} details the resampling and metric recomputation procedure.

\paragraph{Model selection and baselines.}
For all 8B and 32B experiments, validation selection minimizes the equally weighted average of the base-domain and specialist-domain validation losses, $(L_{\mathrm{val,base}}+L_{\mathrm{val,specialist}})/2$. Validation loss is computed on the final assistant response, with preceding context masked out. AMARIS fixes $\beta=1$ in all experiments, retaining base-model embeddings, normalization parameters, and biases.
For the 8B finance and instruction-following settings, we compare against tuned linear interpolation, SLERP, TIES, DARE-TIES, RegMean, LEWIS-TIES, and LEWIS-DARE-TIES with a 75\%/25\% train--validation split and hyperparameters detailed in Table~\ref{tab:sweep_grids_appendix}. For the 32B SEA setting, we compare against tuned linear interpolation and RegMean with an 80\%/20\% train--validation split and hyperparameters detailed in Appendix Table~\ref{tab:sweep_grids_sea_appendix}. Selected hyperparameters are stated in Appendix Tables~\ref{tab:hparams_finance_appendix}--\ref{tab:hparams_sea_appendix}.

Additional untuned 32B SEA baselines are reported in Appendix Table~\ref{tab:sea_untuned}; they are separate from the matched-tuning comparison in Table~\ref{tab:set3_sea_core}.

\subsection{Experimental Settings}

\paragraph{Set 1: Finance (Qwen3-8B + ODA-Fin-RL-8B).}
The base model is Qwen3-8B and the specialist model is ODA-Fin-RL-8B, a finance-specialized model trained from Qwen3-8B \citep{finance}. For base-capability retention, we use 1,000 math questions sampled from the math subset of \citet{Nemotron_Cascade_2}; completions are generated with Qwen3-8B. For specialist transfer, we use 1,000 finance questions from \citet{finance}; completions are generated with ODA-Fin-RL \citep{finance}.

\paragraph{Set 2: Instruction Following (Qwen3-8B + Light-IF-8B).}
The base model is Qwen3-8B and the specialist model is Light-IF-8B, an instruction-following model trained from Qwen3-8B \citep{light}.
For base-capability retention, we again use 1,000 math questions from \citet{Nemotron_Cascade_2} with Qwen3-8B-generated completions. For specialist transfer, we use 1,000 instruction-following prompts from the instruction-following subset of \citet{Nemotron_Cascade_2}, with completions generated by Light-IF-8B \citep{light}.

\paragraph{Set 3: Multilingual SEA (Qwen3-32B + Qwen-SEA-LION-v4-32B-IT).}
The base model is Qwen3-32B and the specialist model is Qwen-SEA-LION-v4-32B-IT by AI Singapore \citep{sealion-qwen3-32b}. For base-capability retention, we use 1,000 math questions from \citet{Nemotron_Cascade_2} with Qwen3-32B-generated completions. For specialist transfer, we use 1,000 documents from \texttt{aisingapore/SEA-Instruct-2602} (SEA-Instruct) \citep{sea_instruct_2602}, covering Indonesian, Vietnamese, Thai, Tamil, and Tagalog. Appendix~\ref{app:sea_calibration} compares this specialist calibration source with \texttt{aisingapore/SEA-PILE-v2} (SEA-PILE) \citep{sea_pile_v2}.

SEA-Instruct calibration uses the dataset's existing full question--answer pairs; we do not generate additional SEA completions.

\subsection{Primary Results}

Tables~\ref{tab:set1_finance}--\ref{tab:set3_sea_core} report all compared methods in each setting. AMARIS attains the highest Core Avg. Preserved in all settings (finance, instruction following, SEA). For the 32B SEA setting, AMARIS also achieves the highest specialist retention among compared merge baselines. Appendix Table~\ref{tab:sea_bootstrap} reports overall 32B differences and confidence intervals; per-language base, specialist, and AMARIS scores and gains over the base are provided in Table~\ref{tab:set3_seahelm_retention}.

\begin{table}[!htbp]
\centering
\small
\caption{Finance merge: Qwen3-8B + ODA-Fin-RL-8B. Brackets show normalized retention percentages; bold and underlining mark the best and second-best merge results.}
\label{tab:set1_finance}
\resizebox{\columnwidth}{!}{%
\begin{tabular}{lcccccc}
\toprule
 & \multicolumn{5}{c}{Core} & \multicolumn{1}{c}{Specialist} \\
\cmidrule(lr){2-6} \cmidrule(l){7-7}
Model & AIME26 & LCB\_v6 & IFEval & GPQA:Diamond & Core Avg. & FPB \\
\midrule
Qwen3-8B & 65.63 [100.00\%] & 48.00 [100.00\%] & 88.17 [100.00\%] & 57.58 [100.00\%] & 64.85 [100.00\%] & 79.23 [94.92\%] \\
ODA-Fin-RL & 58.96 [89.84\%] & 50.86 [105.96\%] & 62.85 [71.28\%] & 48.99 [85.08\%] & 55.42 [85.46\%] & 83.47 [100.00\%] \\
Linear & 63.96 [97.46\%] & 53.14 [110.71\%] & 71.53 [81.13\%] & 45.96 [79.82\%] & 58.65 [90.44\%] & 83.04 [99.48\%] \\
SLERP & 60.00 [91.42\%] & 49.71 [103.56\%] & 74.86 [84.90\%] & 52.02 [90.34\%] & 59.15 [91.21\%] & 82.99 [99.42\%] \\
TIES & 55.83 [85.07\%] & 45.14 [94.04\%] & 68.21 [77.36\%] & 47.98 [83.33\%] & 54.29 [83.72\%] & 82.87 [99.28\%] \\
DARE-TIES & 59.38 [90.48\%] & 46.86 [97.63\%] & 63.03 [71.49\%] & 49.49 [85.95\%] & 54.69 [84.33\%] & \underline{83.40 [99.92\%]} \\
RegMean & 63.54 [96.82\%] & 49.14 [102.38\%] & 75.60 [85.74\%] & 51.52 [89.48\%] & \underline{59.95 [92.44\%]} & 81.89 [98.11\%] \\
LEWIS-TIES & 57.29 [87.29\%] & 49.71 [103.56\%] & 70.79 [80.29\%] & 45.45 [78.93\%] & 55.81 [86.06\%] & 82.38 [98.69\%] \\
LEWIS-DARE-TIES & 62.08 [94.59\%] & 48.00 [100.00\%] & 63.22 [71.70\%] & 49.49 [85.95\%] & 55.70 [85.89\%] & \textbf{84.01 [100.65\%]} \\
AMARIS ($\alpha{=}0.2$) & 62.71 [95.55\%] & 48.57 [101.19\%] & 81.52 [92.46\%] & 54.04 [93.85\%] & \textbf{61.71 [95.16\%]} & 82.11 [98.37\%] \\
\bottomrule
\end{tabular}
}
\end{table}

\begin{table}[!htbp]
\centering
\small
\caption{Instruction-following merge: Qwen3-8B + Light-IF-8B. Brackets show normalized retention percentages; bold and underlining mark the best and second-best merge results.}
\label{tab:set2_if}
\resizebox{\columnwidth}{!}{%
\begin{tabular}{lcccccc}
\toprule
 & \multicolumn{5}{c}{Core} & \multicolumn{1}{c}{Specialist} \\
\cmidrule(lr){2-6} \cmidrule(l){7-7}
Model & AIME26 & LCB\_v6 & IFEval & GPQA:Diamond & Core Avg. & IFBench \\
\midrule
Qwen3-8B & 65.63 [100.00\%] & 48.00 [100.00\%] & 88.17 [100.00\%] & 57.58 [100.00\%] & 64.85 [100.00\%] & 34.33 [76.85\%] \\
Light-IF-8B & 54.58 [83.16\%] & 47.43 [98.81\%] & 90.20 [102.30\%] & 57.07 [99.11\%] & 62.32 [96.10\%] & 44.67 [100.00\%] \\
Linear & 58.33 [88.88\%] & 47.43 [98.81\%] & 90.39 [102.52\%] & 58.59 [101.75\%] & 63.69 [98.21\%] & 41.33 [92.52\%] \\
SLERP & 62.29 [94.91\%] & 47.43 [98.81\%] & 90.94 [103.14\%] & 58.08 [100.87\%] & 64.69 [99.75\%] & 39.00 [87.31\%] \\
TIES & 60.83 [92.69\%] & 49.14 [102.38\%] & 90.02 [102.10\%] & 56.06 [97.36\%] & 64.01 [98.70\%] & 42.67 [95.52\%] \\
DARE-TIES & 53.75 [81.90\%] & 45.71 [95.23\%] & 91.68 [103.98\%] & 58.59 [101.75\%] & 62.43 [96.27\%] & \textbf{45.67 [102.24\%]} \\
RegMean & 63.13 [96.19\%] & 51.43 [107.15\%] & 89.46 [101.46\%] & 57.07 [99.11\%] & \underline{65.27 [100.65\%]} & 43.67 [97.76\%] \\
LEWIS-TIES & 63.33 [96.50\%] & 50.29 [104.77\%] & 90.39 [102.52\%] & 57.07 [99.11\%] & \underline{65.27 [100.65\%]} & \underline{45.00 [100.74\%]} \\
LEWIS-DARE-TIES & 56.67 [86.35\%] & 44.57 [92.85\%] & 90.94 [103.14\%] & 54.04 [93.85\%] & 61.56 [94.93\%] & 43.33 [97.00\%] \\
AMARIS ($\alpha{=}0.2$) & 66.46 [101.26\%] & 49.14 [102.38\%] & 91.87 [104.20\%] & 61.11 [106.13\%] & \textbf{67.15 [103.55\%]} & 40.33 [90.28\%] \\
\bottomrule
\end{tabular}
}
\end{table}

\begin{table}[!htbp]
\centering
\small
\caption{SEA merge: Qwen3-32B + Qwen-SEA-LION-v4-32B-IT, with complete-task SEAHELM evaluation and matched five-configuration tuning. Brackets and highlighting follow the conventions in Section~\ref{sec:experiments}.}
\label{tab:set3_sea_core}
\resizebox{\columnwidth}{!}{%
\begin{tabular}{lcccccc}
\toprule
 & \multicolumn{5}{c}{Core} & \multicolumn{1}{c}{Specialist} \\
\cmidrule(lr){2-6} \cmidrule(l){7-7}
Model & AIME26 & LCB\_v6 & IFEval & GPQA:Diamond & Core Avg. & SEAHELM \\
\midrule
Qwen3-32B & 71.88 [100.00\%] & 58.29 [100.00\%] & 87.62 [100.00\%] & 62.63 [100.00\%] & 70.11 [100.00\%] & 62.69 [92.89\%] \\
Qwen-SEA-LION-v4-32B-IT & 19.17 [26.67\%] & 27.43 [47.06\%] & 87.99 [100.42\%] & 53.54 [85.49\%] & 47.03 [67.08\%] & 67.49 [100.00\%] \\
RegMean & 70.80 [98.50\%] & 57.70 [98.99\%] & 87.40 [99.75\%] & 57.10 [91.17\%] & \underline{68.25 [97.35\%]} & \underline{53.79 [79.70\%]} \\
Linear & 69.17 [96.23\%] & 57.71 [99.00\%] & 87.80 [100.21\%] & 57.58 [91.94\%] & 68.07 [97.09\%] & 52.93 [78.43\%] \\
AMARIS ($\alpha{=}0.001$, $\beta{=}1$) & 71.88 [100.00\%] & 59.40 [101.90\%] & 86.32 [98.52\%] & 60.60 [96.76\%] & \textbf{69.55 [99.20\%]} & \textbf{63.98 [94.80\%]} \\
\bottomrule
\end{tabular}
}
\end{table}

At a high level, results answer \textbf{RQ1} positively: AMARIS repeatedly lands on or near the Pareto frontier. The clearest evidence of specialist transfer occurs in SEA, where AMARIS significantly improves Tamil over the base and overall SEAHELM over the tuned merge baselines.

In the 8B finance setting, AMARIS achieves a core average of 61.71, the highest among all tested merge baselines, and a specialist score of 82.11, close to the best merge baseline score of 84.01.

In the 8B instruction-following setting, AMARIS achieves a core average of 67.15, the highest among all tested merge baselines, but its specialist score of 40.33 is below the best merge baseline score of 45.67. However, AMARIS still achieves higher specialist performance than the base model, and Appendix Table~\ref{tab:bootstrap_8b} shows that this difference is statistically significant.

In the 32B SEA setting, AMARIS achieves a core average of 69.55 and a specialist score of 63.98, both the highest among all tested merge baselines.

\begin{table}[!htbp]
\centering\small
\caption{32B per-language transfer. Differences and paired-bootstrap 95\% CIs are relative to the base. Scores and differences are rounded independently.}
\label{tab:set3_seahelm_retention}
\resizebox{\columnwidth}{!}{%
\begin{tabular}{lrrrrr}
\toprule
Language & Base & AMARIS & Specialist & $\Delta$ vs. base & 95\% CI \\
\midrule
Indonesian & 71.99 & 72.31 & 71.77 & 0.31 & [-0.50, 1.13] \\
Vietnamese & 68.12 & 68.65 & 67.20 & 0.53 & [-0.25, 1.40] \\
Thai & 63.23 & 63.87 & 65.22 & 0.64 & [-0.20, 1.49] \\
Tamil & 41.83 & 46.17 & 63.04 & 4.34 & [2.82, 6.00] \\
Tagalog & 68.27 & 68.91 & 70.20 & 0.64 & [-0.71, 2.03] \\
Overall & 62.69 & 63.98 & 67.49 & 1.29 & [0.80, 1.82] \\
\bottomrule
\end{tabular}}
\end{table}

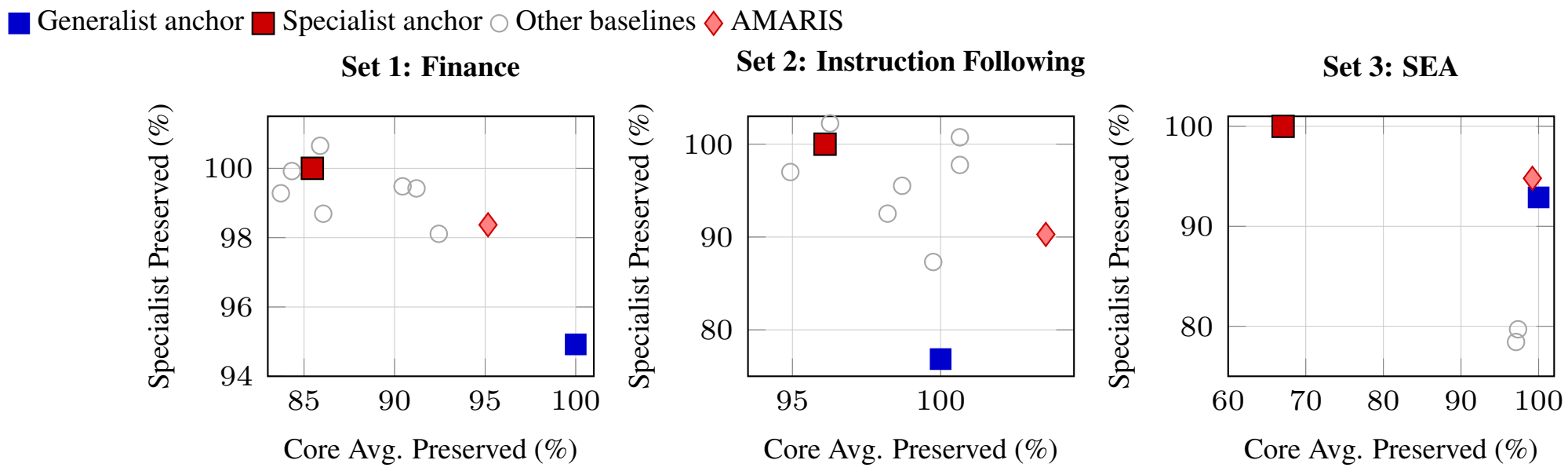
\begin{figure*}[t]
\centering
\begin{tikzpicture}
\begin{groupplot}[
  group style={group size=3 by 1, horizontal sep=1.3cm},
  width=0.31\textwidth,
  height=0.27\textwidth,
  grid=both,
  grid style={line width=.1pt, draw=gray!20},
  major grid style={line width=.2pt,draw=gray!40},
  tick label style={font=\scriptsize},
  label style={font=\scriptsize},
  title style={font=\scriptsize\bfseries},
  xlabel={Core Avg. Preserved (\%)},
  ylabel={Specialist Preserved (\%)},
  legend style={font=\scriptsize, draw=none, fill=none, at={(0.5,1.24)}, anchor=south, legend columns=4}
]

\nextgroupplot[
title={Set 1: Finance},
xmin=83,
xmax=101,
ymin=94,
ymax=101.5
]
\addplot+[only marks, mark=square*, mark size=2.4pt, color=blue!80!black] coordinates {
  (100.00,94.92)
};
\addlegendentry{Generalist anchor}
\addplot+[only marks, mark=square*, mark size=2.6pt, color=black] coordinates {
  (85.46,100.00)
};
\addlegendentry{Specialist anchor}
\addplot+[only marks, mark=o, mark size=2.0pt, color=gray!70] coordinates {
  (90.44,99.48)
  (91.21,99.42)
  (83.72,99.28)
  (84.33,99.92)
  (92.44,98.11)
  (86.06,98.69)
  (85.89,100.65)
};
\addlegendentry{Other baselines}
\addplot+[only marks, mark=diamond*, mark size=2.8pt, color=red!80!black, fill=red!50] coordinates {(95.16,98.37)};
\addlegendentry{AMARIS}

\nextgroupplot[
title={Set 2: Instruction Following},
xmin=93.5,
xmax=104.5,
ymin=75,
ymax=103
]
\addplot+[only marks, mark=square*, mark size=2.4pt, color=blue!80!black] coordinates {
  (100.00,76.85)
};
\addplot+[only marks, mark=square*, mark size=2.6pt, color=black] coordinates {
  (96.10,100.00)
};
\addplot+[only marks, mark=o, mark size=2.0pt, color=gray!70] coordinates {
  (98.21,92.52)
  (99.75,87.31)
  (98.70,95.52)
  (96.27,102.24)
  (100.65,97.76)
  (100.65,100.74)
  (94.93,97.00)
};
\addplot+[only marks, mark=diamond*, mark size=2.8pt, color=red!80!black, fill=red!50] coordinates {(103.55,90.28)};

\nextgroupplot[
title={Set 3: SEA},
xmin=60,
xmax=102,
ymin=75,
ymax=101
]
\addplot+[only marks, mark=square*, mark size=2.4pt, color=blue!80!black] coordinates {
  (100.00,92.89)
};
\addplot+[only marks, mark=square*, mark size=2.6pt, color=black] coordinates {
  (67.08,100.00)
};
\addplot+[only marks, mark=o, mark size=2.0pt, color=gray!70] coordinates {
  (97.35,79.70)
  (97.09,78.43)
};
\addplot+[only marks, mark=diamond*, mark size=2.8pt, color=red!80!black, fill=red!50] coordinates {(99.20,94.80)};

\end{groupplot}
\end{tikzpicture}
\caption{\textbf{Core--specialist operating points.} Core Avg. Preserved versus specialist-normalized score for the three settings, using point estimates from Tables~\ref{tab:set1_finance}--\ref{tab:set3_sea_core}.}
\label{fig:pareto_three_sets_tikz}
\end{figure*}

\subsection{Statistical Uncertainty}
\label{sec:statistical_uncertainty}

The paired-bootstrap comparisons in Table~\ref{tab:bootstrap_main} qualify the point-estimate rankings in Tables~\ref{tab:set1_finance}--\ref{tab:set3_sea_core}. We compare AMARIS with the highest-scoring specialist merge baseline and RegMean which is also a data conditioned merging method to AMARIS in each setting to assess statistical significance.

\begin{table}[!htbp]
\centering\small
\caption{Paired-bootstrap comparisons with selected merge baselines. Differences are AMARIS minus the comparator. Specialist metrics are FPB for finance, IFBench for instruction following (IF), and overall SEAHELM for SEA.}
\label{tab:bootstrap_main}
\resizebox{\columnwidth}{!}{%
\begin{tabular}{llrrrr}
\toprule
Setting & Comparator & $\Delta$ Core Avg. & 95\% CI & $\Delta$ Specialist & 95\% CI \\
\midrule
8B Finance & LEWIS-DARE-TIES & 6.01 & [3.07, 8.90] & -1.89 & [-2.69, -1.09] \\
8B Finance & RegMean & 1.76 & [-0.62, 4.13] & 0.23 & [-0.50, 0.95] \\
\midrule
8B IF & DARE-TIES & 4.71 & [2.08, 7.38] & -5.34 & [-10.67, 0.00] \\
8B IF & RegMean & 1.87 & [-0.41, 4.17] & -3.33 & [-8.33, 1.34] \\
\midrule
32B SEA & Linear & 1.48 & [-0.80, 3.87] & 11.05 & [9.99, 12.04] \\
32B SEA & RegMean & 1.30 & [-1.11, 3.80] & 10.19 & [9.15, 11.15] \\
\bottomrule
\end{tabular}}
\end{table}

In 8B finance, AMARIS significantly improves Core Avg. over LEWIS-DARE-TIES (+6.01 points), but significantly reduces FPB ($-1.89$ points). Against RegMean, neither the Core Avg. nor FPB difference is statistically significant.

In 8B instruction following, AMARIS significantly improves Core Avg. over DARE-TIES (+4.71 points), while the IFBench difference ($-5.34$ points) is not statistically significant. Neither difference against RegMean is statistically significant.

In 32B SEA, AMARIS significantly improves overall SEAHELM over RegMean (+10.19 points) and Linear (+11.05 points), while neither Core Avg. difference is statistically significant. Per-language comparisons with the base show a significant improvement only for Tamil (+4.34 points; Table~\ref{tab:set3_seahelm_retention}).

Table~\ref{tab:bootstrap_main} reports the 95\% confidence intervals; Appendix~\ref{app:bootstrap} provides the full comparisons and resampling procedure.

\section{Trade-off Analysis}
\label{sec:analysis_summary}

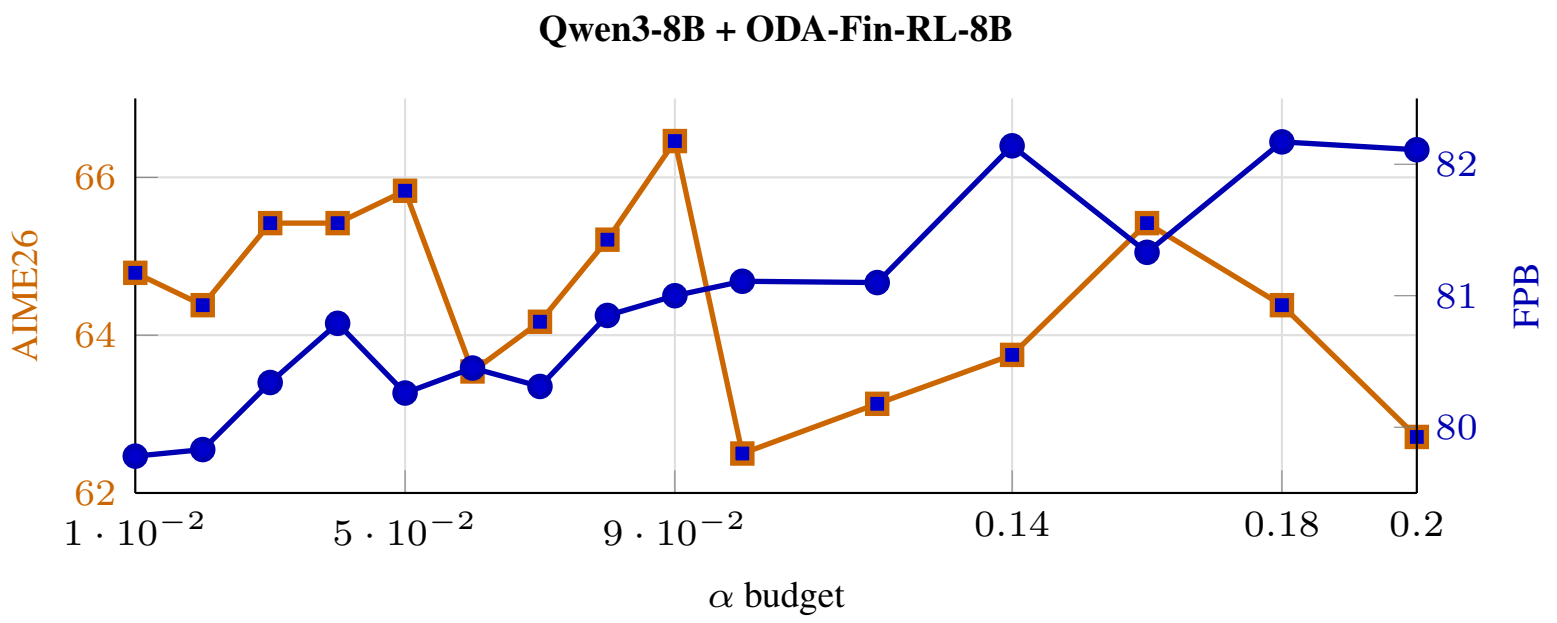
\begin{figure*}[!htbp]
\centering
\begin{tikzpicture}
\begin{axis}[
  name=leftpanel,
  width=0.72\textwidth,
  height=0.30\textwidth,
  xmin=0.01,
  xmax=0.20,
  ymin=62.0,
  ymax=67.0,
  xtick={0.01,0.05,0.09,0.14,0.18,0.20},
  xlabel={$\alpha$ budget},
  ylabel={AIME26},
  ylabel style={color=orange!80!black},
  yticklabel style={color=orange!80!black},
  tick label style={font=\scriptsize},
  label style={font=\scriptsize},
  title style={font=\scriptsize\bfseries},
  title={Qwen3-8B + ODA-Fin-RL-8B},
  axis y line*=left,
  axis x line*=bottom,
  grid=major,
  grid style={draw=gray!25}
]
\addplot+[color=orange!80!black, mark=square*, mark size=1.8pt, line width=1.0pt] coordinates {
  (0.01,64.79)
  (0.02,64.38)
  (0.03,65.42)
  (0.04,65.42)
  (0.05,65.83)
  (0.06,63.54)
  (0.07,64.17)
  (0.08,65.21)
  (0.09,66.46)
  (0.10,62.50)
  (0.12,63.13)
  (0.14,63.75)
  (0.16,65.42)
  (0.18,64.38)
  (0.20,62.71)
};
\end{axis}

\begin{axis}[
  at={(leftpanel.south west)},
  anchor=south west,
  width=0.72\textwidth,
  height=0.30\textwidth,
  xmin=0.01,
  xmax=0.20,
  ymin=79.5,
  ymax=82.5,
  ylabel={FPB},
  ylabel style={color=blue!70!black},
  yticklabel style={color=blue!70!black},
  tick label style={font=\scriptsize},
  label style={font=\scriptsize},
  axis y line*=right,
  axis x line=none
]
\addplot+[color=blue!70!black, mark=*, mark size=1.9pt, line width=1.0pt] coordinates {
  (0.01,79.78)
  (0.02,79.83)
  (0.03,80.34)
  (0.04,80.79)
  (0.05,80.26)
  (0.06,80.45)
  (0.07,80.31)
  (0.08,80.85)
  (0.09,81.00)
  (0.10,81.11)
  (0.12,81.10)
  (0.14,82.14)
  (0.16,81.33)
  (0.18,82.17)
  (0.20,82.11)
};
\end{axis}
\end{tikzpicture}
\caption{\textbf{Finance $\alpha$ sweep.} AIME26 (left axis) and FPB (right axis) for Qwen3-8B + ODA-Fin-RL-8B; their highest observed scores occur at different budgets.}
\label{fig:alpha_sensitivity_tikz}
\end{figure*}

Figure~\ref{fig:pareto_three_sets_tikz} shows descriptive operating points across settings. Figure~\ref{fig:alpha_sensitivity_tikz} shows that in finance, FPB peaks at $\alpha=0.18$ while AIME26 peaks at $\alpha=0.09$, indicating non-aligned optima. This sweep partially addresses \textbf{RQ2}: changing $\alpha$ affects the observed core--specialist trade-off in finance, and the two benchmarks favor different budgets.

\section{Limitations}
\label{sec:limitations}

AMARIS has practical constraints that should be considered when interpreting results and planning deployment.

First, the method depends on calibration activations to estimate $C_1$ and $C_2$. If calibration sets are small, distribution-shifted, or noisy relative to deployment traffic, the selected subspace may reduce transfer reliability.

Second, although AMARIS is training-free, outcomes still depend on design choices, including preservation budget $\alpha$, interpolation strength $\beta$, and the set of projected layers.

Third, our experiments focus on pairwise merges with shared architecture and tokenizer. The current formulation does not directly address heterogeneous checkpoints, multi-model composition with conflicting specialist objectives, or iterative merge pipelines where interference may accumulate over stages.

\section{Conclusion}

This paper introduced AMARIS, a training-free merging framework that formulates specialist inheritance as budgeted subspace selection over activation covariances. By ranking and selecting low-interference directions under an explicit preservation budget, AMARIS transfers specialist updates while controlling perturbation to base-model behavior.

Across the evaluated finance, instruction-following, and multilingual SEA settings, AMARIS yields competitive specialist--generalist operating points. The SEA evidence supports a narrow conclusion: it preserves much of the base behavior, avoids substantial degradation relative to matched-tuned RegMean and Linear Merge, and partially transfers the specialist Tamil gain; it does not establish full specialist inheritance or general dominance over all baselines.

Taken together, these results support subspace-constrained merging as a practical route to compositional capability integration without additional training and provide evidence for \textbf{RQ3} across the three specialization regimes. Promising next steps include multi-specialist composition, heterogeneous checkpoint alignment, and more automated budget selection.

\section*{Acknowledgements}
This paper was fully funded and supported by DSO National Laboratories.

\section*{Competing Interests}
The authors declare no competing interests.

\clearpage

\bibliographystyle{plainnat}
\bibliography{references}

\clearpage


\appendix

\section{Paired-Bootstrap Comparisons}
\label{app:bootstrap}

Tables~\ref{tab:bootstrap_8b} and~\ref{tab:sea_bootstrap} report the 8B and 32B comparisons, respectively. All intervals resample evaluation examples for fixed checkpoints; they do not quantify seed, decoding or calibration variability. We use 10,000 paired resamples and percentile 95\% intervals.

\paragraph{Resampling procedure.}
We resample questions with replacement, using the same sampled question IDs for AMARIS and its comparator in each replicate. For benchmarks with multiple generations per question, we first average the generation scores into one question-level score; individual generations are not resampled separately. Each replicate recomputes the benchmark scores and their core average. For SEAHELM, questions are resampled separately within each language--task stratum, and task scores are reaggregated into language scores and the overall score in each replicate. For FPB, weighted F1 is recomputed from the sampled gold-label and prediction pairs rather than averaged from question-level F1 values. We compute AMARIS-minus-comparator differences within each replicate and report their percentile 95\% intervals.

\begin{table}[t]
\centering\small
\caption{8B paired-bootstrap comparisons. Differences are AMARIS minus the comparator, in absolute score units. Specialist metrics are FPB for Finance and IFBench for IF.}
\label{tab:bootstrap_8b}
\resizebox{\columnwidth}{!}{%
\begin{tabular}{llrrrr}
\toprule
Setting & Comparator & $\Delta$ Core & 95\% CI & $\Delta$ Specialist & 95\% CI \\
\midrule
Finance & LEWIS-DARE-TIES & 6.01 & [3.07, 8.90] & -1.89 & [-2.69, -1.09] \\
Finance & Linear & 3.06 & [0.29, 5.80] & -0.92 & [-1.70, -0.16] \\
Finance & RegMean & 1.76 & [-0.62, 4.13] & 0.23 & [-0.50, 0.95] \\
Finance & Base & -3.13 & [-5.66, -0.63] & 2.88 & [1.99, 3.79] \\
Finance & Specialist & 6.30 & [3.50, 9.03] & -1.35 & [-2.11, -0.55] \\
IF & DARE-TIES & 4.71 & [2.08, 7.38] & -5.34 & [-10.67, 0.00] \\
IF & Linear & 3.46 & [0.94, 5.95] & -1.00 & [-6.00, 3.67] \\
IF & RegMean & 1.87 & [-0.41, 4.17] & -3.33 & [-8.33, 1.34] \\
IF & Base & 2.30 & [-0.06, 4.70] & 6.00 & [0.67, 11.00] \\
IF & Specialist & 4.82 & [2.12, 7.51] & -4.33 & [-9.67, 1.00] \\
\bottomrule
\end{tabular}}
\end{table}

\begin{table}[t]
\centering\small
\caption{32B core-average and overall SEAHELM paired-bootstrap comparisons. Differences are AMARIS minus the comparator; intervals are percentile 95\% intervals.}
\label{tab:sea_bootstrap}
\resizebox{\columnwidth}{!}{%
\begin{tabular}{lrrrr}
\toprule
Comparator & $\Delta$ Core & 95\% CI & $\Delta$ SEAHELM & 95\% CI \\
\midrule
Base & -0.56 & [-2.86, 1.80] & 1.29 & [0.80, 1.82] \\
Specialist & 22.52 & [18.27, 26.67] & -3.51 & [-4.33, -2.66] \\
RegMean & 1.30 & [-1.11, 3.80] & 10.19 & [9.15, 11.15] \\
Linear & 1.48 & [-0.80, 3.87] & 11.05 & [9.99, 12.04] \\
\bottomrule
\end{tabular}}
\end{table}

\section{Untuned 32B SEA Baselines}
\label{app:sea_untuned}

Table~\ref{tab:sea_untuned} reports additional default-configuration baselines without validation tuning. These point estimates provide context for the main comparison; they do not establish relative performance under matched tuning budgets. Their SEAHELM scores range from 53.91 to 54.85, compared with 63.98 for tuned AMARIS. Tamil scores range from 16.02 to 16.94, below the base model's 41.83.

\begin{table}[!htbp]
\centering
\small
\caption{Untuned 32B SEA baselines, with source checkpoints shown for reference. Core and specialist columns, bracketed retention percentages, and highlighting follow Tables~\ref{tab:set1_finance}--\ref{tab:set3_sea_core}; best and second-best results are identified among the untuned merging methods in this table. The per-language panel reports unnormalized point estimates.}
\label{tab:sea_untuned}
\resizebox{\columnwidth}{!}{%
\begin{tabular}{lcccccc}
\toprule
 & \multicolumn{5}{c}{Core} & \multicolumn{1}{c}{Specialist} \\
\cmidrule(lr){2-6} \cmidrule(l){7-7}
Model & AIME26 & LCB\_v6 & IFEval & GPQA:Diamond & Core Avg. & SEAHELM \\
\midrule
Qwen3-32B & 71.88 [100.00\%] & 58.29 [100.00\%] & 87.62 [100.00\%] & 62.63 [100.00\%] & 70.11 [100.00\%] & 62.69 [92.89\%] \\
Qwen-SEA-LION-v4-32B-IT & 19.17 [26.67\%] & 27.43 [47.06\%] & 87.99 [100.42\%] & 53.54 [85.49\%] & 47.03 [67.08\%] & 67.49 [100.00\%] \\
TIES & 64.79 [90.14\%] & 60.57 [103.91\%] & 86.69 [98.94\%] & 56.57 [90.32\%] & \underline{67.16 [95.79\%]} & 54.62 [80.93\%] \\
DARE-TIES & 58.75 [81.73\%] & 49.71 [85.28\%] & 87.25 [99.58\%] & 60.10 [95.96\%] & 63.95 [91.21\%] & 54.49 [80.74\%] \\
LEWIS-TIES & 62.71 [87.24\%] & 54.86 [94.12\%] & 85.03 [97.04\%] & 61.11 [97.57\%] & 65.93 [94.04\%] & \underline{54.78 [81.17\%]} \\
LEWIS-DARE-TIES & 57.08 [79.41\%] & 52.57 [90.19\%] & 87.25 [99.58\%] & 60.61 [96.77\%] & 64.38 [91.83\%] & \textbf{54.85 [81.27\%]} \\
SLERP & 68.96 [95.94\%] & 58.29 [100.00\%] & 86.32 [98.52\%] & 61.11 [97.57\%] & \textbf{68.67 [97.95\%]} & 53.91 [79.88\%] \\
\midrule
\multicolumn{7}{l}{\textit{SEAHELM by language}} \\
Model & Indonesian & Vietnamese & Thai & Tamil & Tagalog & Overall \\
\midrule
TIES & 68.51 & 63.62 & 59.72 & 16.12 & 65.12 & 54.62 \\
DARE-TIES & 67.42 & 63.15 & 59.45 & 16.94 & 65.51 & 54.49 \\
LEWIS-TIES & 68.16 & 64.08 & 59.58 & 16.31 & 65.77 & 54.78 \\
LEWIS-DARE-TIES & 68.45 & 64.03 & 59.70 & 16.33 & 65.77 & 54.85 \\
SLERP & 67.13 & 63.90 & 58.29 & 16.02 & 64.20 & 53.91 \\
\bottomrule
\end{tabular}}
\end{table}

\section{SEA 32B Calibration-Source Comparison}
\label{app:sea_calibration}

We selected $\alpha=0.001$ and $\beta=1$ using validation-loss tuning with SEA-Instruct \citep{sea_instruct_2602}, which supplies the main-text results. To assess sensitivity to the specialist calibration source, we reran AMARIS using SEA-PILE \citep{sea_pile_v2} at the same $\alpha$ and $\beta$, without retuning. Both calibration-source conditions use the same complete-task SEAHELM evaluation, multi-turn judge, eigensolver, and all remaining merge and evaluation settings; only the specialist calibration source changes. Table~\ref{tab:sea_calibration} reports the per-language SEAHELM scores and the four core benchmarks, together with their aggregate scores.

\begin{table}[!htbp]
\centering\small
\caption{SEA 32B calibration-source comparison at $\alpha=0.001$, $\beta=1$. Higher is better for all scores. Differences are SEA-PILE minus SEA-Instruct in absolute score units, calculated from the displayed scores. Core Avg. is the arithmetic mean of the four core benchmark scores.}
\label{tab:sea_calibration}
\resizebox{\columnwidth}{!}{%
\begin{tabular}{lrrr}
\toprule
Metric & SEA-Instruct & SEA-PILE & $\Delta$ \\
\midrule
\multicolumn{4}{l}{\textit{SEAHELM}} \\
Indonesian & 72.31 & 72.45 & +0.14 \\
Vietnamese & 68.65 & 68.72 & +0.07 \\
Thai & 63.87 & 63.92 & +0.05 \\
Tamil & 46.17 & 46.80 & +0.63 \\
Tagalog & 68.91 & 69.27 & +0.36 \\
\textbf{SEAHELM Overall} & \textbf{63.98} & \textbf{64.23} & \textbf{+0.25} \\
\midrule
\multicolumn{4}{l}{\textit{Core benchmarks}} \\
AIME26 & 71.88 & 73.75 & +1.87 \\
LCB\_v6 & 59.40 & 58.86 & -0.54 \\
IFEval & 86.32 & 84.84 & -1.48 \\
GPQA:Diamond & 60.60 & 59.60 & -1.00 \\
\textbf{Core Avg.} & \textbf{69.55} & \textbf{69.26} & \textbf{-0.29} \\
\bottomrule
\end{tabular}}
\end{table}

SEA-PILE improves SEAHELM overall by 0.25 points, with higher scores in all five languages, while Core Avg. decreases by 0.29 points. The two calibration sources yield similar point estimates: SEA-PILE scores slightly higher on SEAHELM, while SEA-Instruct scores slightly higher on the core benchmarks.

\section{Efficiency and Selected Ranks}

Table~\ref{tab:runtime} reports pipeline runtime and peak GPU memory at 8B and 32B scales.

\begin{table}[t]
\centering\small
\caption{Measured pipeline runtime and peak GPU memory. Basis construction and greedy selection were timed jointly.}
\label{tab:runtime}
\resizebox{\columnwidth}{!}{%
\begin{tabular}{lrrrrrl}
\toprule
Scale & Cov. (min) & Basis/greedy (min) & Merge/save (min) & Total (min) & Peak (GiB) & Hardware \\
\midrule
8B & 13 & 1 & 1 & 15 & 49.33 & 1 $\times$ H200 \\
32B & 92 & 11 & 2 & 105 & 163.34 & 2 $\times$ H200 \\
\bottomrule
\end{tabular}}
\end{table}

\paragraph{Accepted ranks.} Table~\ref{tab:sea_rank_analysis} summarizes the reported 32B SEA ranks by module and layer group at $\alpha=0.001$, $\beta=1$, with a per-module candidate cap of 1024. Accepted ranks result from layerwise budgeted selection; the candidate cap is not the selected rank or a shared rank budget across layers.

\begin{table}[!htbp]
\centering\small
\caption{Mean accepted ranks across each 16-layer group, rounded to the nearest integer, for the 32B SEA checkpoint at $\alpha=0.001$, $\beta=1$ (candidate cap: 1024). Layer numbering is one-based.}
\label{tab:sea_rank_analysis}
\resizebox{\columnwidth}{!}{%
\begin{tabular}{lrrrr}
\toprule
Module & Layers 1--16 & Layers 17--32 & Layers 33--48 & Layers 49--64 \\
\midrule
\texttt{q\_proj} & 634 & 119 & 66 & 16 \\
\texttt{k\_proj} & 571 & 117 & 48 & 12 \\
\texttt{v\_proj} & 550 & 100 & 35 & 11 \\
\texttt{o\_proj} & 738 & 631 & 428 & 260 \\
\texttt{gate\_proj} & 531 & 15 & 13 & 6 \\
\texttt{up\_proj} & 527 & 14 & 12 & 7 \\
\texttt{down\_proj} & 624 & 203 & 220 & 315 \\
\bottomrule
\end{tabular}}
\end{table}

The earliest layer group (layers 1--16) has the highest reported ranks for every module, ranging from 527 to 738. Ranks generally decrease with depth, but the pattern differs by projection. Attention output projections (\texttt{o\_proj}) retain relatively high ranks throughout the network (738, 631, 428, and 260 across successive groups), whereas attention query, key, and value projections fall to 11--16 in the final group. Later MLP input projections (\texttt{gate\_proj} and \texttt{up\_proj}) use much smaller ranks, only 6--15 across layers 17--64. The MLP output projection (\texttt{down\_proj}) is also lower than in the earliest group, but retains substantially larger ranks of 203--315 and increases again in the final group. Overall, the selected transfer subspaces are largest in early layers, with substantial depth-dependent differences between projection types.

\section{Reproducibility Details}
\label{app:reproducibility}

\paragraph{Eigensolver and candidate caps.}
For the reported experiments, we use LOBPCG to approximate leading generalized eigenpairs, retaining up to 512 candidate directions per targeted weight for 8B models and 1024 for 32B models before budgeted greedy selection. These caps limit the candidate pool, not the final accepted rank.

\paragraph{Evaluation modes.}
AIME26, LCB\_v6, GPQA-Diamond, IFBench, and FPB use thinking mode; IFEval and SEAHELM use non-thinking mode. These modes apply to all evaluated checkpoints; decoding settings are listed in Table~\ref{tab:decoding}.

\paragraph{Generated vs prompts-only calibration.}
Calibration uses teacher forcing on complete prompt--response pairs. For the 8B experiments and the 32B base calibration set, responses are model-generated as described in Section~\ref{sec:experiments}. For SEA-Instruct, we use the existing full question--answer pairs without generating additional completions.

\paragraph{Targeted weights and numerical fallback.}
Projection applies to the weights of targeted \texttt{nn.Linear} modules: attention projections (\texttt{q\_proj}, \texttt{k\_proj}, \texttt{v\_proj}, \texttt{o\_proj}), MLP projections (\texttt{up\_proj}, \texttt{gate\_proj}, \texttt{down\_proj}), and \texttt{lm\_head}. The output head uses the same projected-delta rule unless excluded by configuration or assigned a numerical fallback. All remaining parameters use $\beta$ interpolation, including embeddings, RMSNorm parameters, and linear biases.

\begin{table}[t]
\centering\small
\caption{Decoding and evaluation settings. Samples denotes generations per question; SEA token limits vary by task.}
\label{tab:decoding}
\resizebox{\columnwidth}{!}{%
\begin{tabular}{lrrrrrrl}
\toprule
Benchmark & Temp. & Top-$p$ & Top-$k$ & Max new tokens & Seed & Samples & Metric \\
\midrule
AIME26 & 0.6 & 0.95 & 20 & 32768 & 1234 & 16 & pass@1 mean of 16 \\
LCB\_v6 & 0.6 & 0.95 & 20 & 32768 & 1234 & 16 & pass@1 unit tests mean of 16 \\
GPQA-Diamond & 0.6 & 0.95 & 20 & 32768 & 1234 & 8 & pass@1 mean of 8 \\
IFEval & 0.6 & 0.95 & 20 & 32768 & 1234 & 1 & prompt-level loose accuracy \\
IFBench & 0.6 & 0.95 & 20 & 32768 & 1234 & 1 & prompt-level loose accuracy pass@1 \\
FPB & 0.6 & 0.95 & 20 & 32768 & 1234 & 1 & weighted F1 \\
SEAHELM & 0.7 & 0.8 & 20 & 16--1024 per task & 1234 & 1 & official per-task metrics \\
\bottomrule
\end{tabular}}
\end{table}

\section{Derivation of the Formulation Objective}
\label{app:setup_derivation}

The formulation in Section~\ref{sec:setup} is motivated by two output-space objectives at one layer:
\[
\|Y_{\text{merge}}(X_1)-Y_1(X_1)\|_F^2,
\qquad
\|Y_{\text{merge}}(X_2)-Y_2(X_2)\|_F^2.
\]
With $W_{\text{merge}}=W_1+\Delta P$ and $P=P^\top=P^2$,
\[
Y_{\text{merge}}(X)-Y_1(X)=XP\Delta^\top,
\qquad
Y_{\text{merge}}(X)-Y_2(X)=-X(I-P)\Delta^\top.
\]
By $\|AB\|_F\le\|A\|_F\|B\|_2$,
\[
\|Y_{\text{merge}}(X_1)-Y_1(X_1)\|_F
=\|X_1P\Delta^\top\|_F
\le\|X_1P\|_F\|\Delta\|_2,
\]
\[
\|Y_{\text{merge}}(X_2)-Y_2(X_2)\|_F
=\|X_2(I-P)\Delta^\top\|_F
\le\|X_2(I-P)\|_F\|\Delta\|_2.
\]
For a fixed layer update $\Delta$, these inequalities bound output discrepancies by routing energies $\|X_1P\|_F^2$ and $\|X_2(I-P)\|_F^2$. Reducing these energies tightens the bounds, but need not minimize the exact output errors. This motivates covariance-only routing terms for geometric interpretation and direction scoring:
\[
\frac{1}{N_1}\|X_1P\|_F^2
=\frac{1}{N_1}\mathrm{tr}(PX_1^\top X_1P)
=\frac{1}{N_1}\mathrm{tr}(PX_1^\top X_1)
=\mathrm{tr}(PC_1),
\]
where idempotence ($P^2=P$) and cyclic trace are used. Likewise,
\[
\frac{1}{N_2}\|X_2(I-P)\|_F^2
=\mathrm{tr}((I-P)C_2)
=\mathrm{tr}(C_2)-\mathrm{tr}(PC_2).
\]
Thus minimizing specialist routing energy is equivalent to maximizing captured activation energy $\mathrm{tr}(PC_2)$. This is a proxy for specialist output preservation: the exact mean squared output error is $\mathrm{tr}((I-P)C_2(I-P)G)$, where $G=\Delta^\top\Delta$. Similarly, constraining $\mathrm{tr}(PC_1)$ bounds base routing energy rather than measuring exact base output perturbation. Normalizing this routing quantity by $\mathrm{tr}(C_1)$ gives
\[
\frac{\mathrm{tr}(PC_1)}{\mathrm{tr}(C_1)}\le\alpha.
\]
If $\Delta$ is retained explicitly in covariance/Gram form, the exact base term is
\[
\frac{1}{N_1}\|X_1P\Delta^\top\|_F^2
=\frac{1}{N_1}\mathrm{tr}(\Delta P X_1^\top X_1 P \Delta^\top)
=\mathrm{tr}(PC_1P\Delta^\top\Delta)
=\mathrm{tr}(PC_1PG),
\]
where $G=\Delta^\top\Delta$. This yields the exact normalized budget
\[
\mathrm{tr}(PC_1PG)\le \alpha\,\mathrm{tr}(C_1G).
\]
In the final implementation, ridge stabilization replaces $C_1$ by $B=C_1+\eta I$ in both the generalized eigensystem and the Gram-form feasibility budget; the unregularized $C_1$ formulation above remains the conceptual preservation objective.
For an orthogonal projector, this regularized cost satisfies
\[
\mathrm{tr}(PBPG)=L_{\mathrm{base}}(P)+\eta\,\mathrm{tr}(PG),
\qquad
\mathrm{tr}(BG)=\mathrm{tr}(C_1G)+\eta\,\mathrm{tr}(G).
\]
Thus feasibility under the regularized budget does not by itself imply
$L_{\mathrm{base}}(P)\le\alpha\,\mathrm{tr}(C_1G)$.

\section{Selected Hyperparameters}
\label{app:selected_hparams}

\subsection{Validation-Tuned Hyperparameter Sweep Grids}
\label{app:sweep_grids}

We tune methods over fixed candidate grids and select by the equally weighted average of base-domain and specialist-domain validation losses at both scales. AMARIS fixes $\beta=1$ for all 8B and 32B experiments. Table~\ref{tab:sweep_grids_appendix} lists the 8B finance and instruction-following grids; Table~\ref{tab:sweep_grids_sea_appendix} lists the matched 32B SEA grids.

\begin{table}[t]
\centering
\small
\caption{Hyperparameter sweep grids used for validation-loss tuning (finance and instruction following).}
\label{tab:sweep_grids_appendix}
\begin{tabular}{@{}p{0.28\columnwidth}p{\dimexpr0.72\columnwidth-2\tabcolsep\relax}@{}}
\toprule
Method & Sweep grid \\
\midrule
Linear ($w$) & $\{0.1, 0.2, 0.3, 0.4, 0.5, 0.6, 0.7, 0.8, 0.9, 1.0\}$ \\
SLERP ($t$) & $\{0.1, 0.2, 0.3, 0.4, 0.5, 0.6, 0.7, 0.8, 0.95\}$ \\
TIES ($d$-$s$) & $\{0.2$-$0.7,\; 0.2$-$1.0,\; 0.3$-$0.7,\; 0.3$-$1.0,\; 0.5$-$0.7,\; 0.5$-$1.0,\; 0.5$-$1.3,\; 0.7$-$0.7,\; 0.7$-$1.0,\; 0.9$-$1.0\}$ \\
DARE-TIES ($d$-$s$) & $\{0.05$-$1.0,\; 0.1$-$1.0,\; 0.1$-$1.5,\; 0.2$-$1.0,\; 0.2$-$1.5,\; 0.3$-$0.7,\; 0.3$-$1.0,\; 0.5$-$1.0,\; 0.5$-$1.5,\; 0.7$-$1.0\}$ \\
LEWIS (for TIES and DARE-TIES, $g$-$e$) & $\{0.1$-$0.4,\; 0.1$-$0.6,\; 0.2$-$0.5,\; 0.2$-$0.7,\; 0.3$-$0.6,\; 0.3$-$0.8,\; 0.4$-$0.7,\; 0.4$-$0.9,\; 0.5$-$0.8,\; 0.5$-$0.95\}$ \\
RegMean ($w$) & $\{0.1, 0.2, 0.3, 0.4, 0.5, 0.6, 0.7, 0.8, 0.9, 1.0\}$ \\
AMARIS ($\alpha$) & $\{0.02, 0.04, 0.06, 0.08, 0.10, 0.12, 0.14, 0.16, 0.18, 0.20\}$ \\
\bottomrule
\end{tabular}
\end{table}

\begin{table}[t]
\centering
\small
\caption{Hyperparameter sweep grids used for validation-loss tuning (SEA). AMARIS fixes $\beta=1$.}
\label{tab:sweep_grids_sea_appendix}
\begin{tabular}{ll}
\toprule
Method & Sweep grid \\
\midrule
Linear ($w$) & $\{0.1, 0.3, 0.5, 0.7, 0.9\}$ \\
RegMean ($w$) & $\{0.2, 0.4, 0.6, 0.8, 1.0\}$ \\
AMARIS ($\alpha$) & $\{0.001, 0.005, 0.01, 0.015, 0.02\}$ \\
\bottomrule
\end{tabular}
\end{table}

\paragraph{Symbol legend for baseline hyperparameters.}
$w$ denotes the linear mixing coefficient for Linear and RegMean, and $t$ denotes the spherical interpolation parameter for SLERP. For TIES and DARE-TIES, $d$ controls task-vector density and $s$ is the delta-reconstruction scaling coefficient. For LEWIS-guided variants, $g$ and $e$ denote the lower and upper task-vector sparsity bounds, respectively, corresponding to $\gamma$ and $\epsilon$ in \citet{lewis}. For RegMean, we fix the non-diagonal multiplier at the default specified by \citet{jin2023dataless} and tune only $w$. AMARIS uses $\alpha$ as the preservation budget described in Section~\ref{sec:method}.

Tables~\ref{tab:hparams_finance_appendix}, \ref{tab:hparams_if_appendix}, and~\ref{tab:hparams_sea_appendix} report the selected hyperparameters for each merge method.

\begin{table}[t]
\centering
\small
\caption{Selected hyperparameters for the finance setting (Qwen3-8B + ODA-Fin-RL-8B), tuned via a 75\%/25\% train--validation split.}
\label{tab:hparams_finance_appendix}
\begin{tabular}{lccccccc}
\toprule
Method & $w$ & $t$ & $d$ & $s$ & $g$ & $e$ & $\alpha$ \\
\midrule
Linear & 0.8 & -- & -- & -- & -- & -- & -- \\
SLERP & -- & 0.8 & -- & -- & -- & -- & -- \\
TIES & -- & -- & 0.5 & 1.0 & -- & -- & -- \\
DARE-TIES & -- & -- & 0.7 & 1.0 & -- & -- & -- \\
RegMean & 1.0 & -- & -- & -- & -- & -- & -- \\
LEWIS-TIES & -- & -- & -- & -- & 0.4 & 0.7 & -- \\
LEWIS-DARE-TIES & -- & -- & -- & -- & 0.5 & 0.95 & -- \\
AMARIS & -- & -- & -- & -- & -- & -- & 0.2 \\
\bottomrule
\end{tabular}
\end{table}

\begin{table}[t]
\centering
\small
\caption{Selected hyperparameters for the instruction-following setting (Qwen3-8B + Light-IF-8B), tuned via a 75\%/25\% train--validation split.}
\label{tab:hparams_if_appendix}
\begin{tabular}{lccccccc}
\toprule
Method & $w$ & $t$ & $d$ & $s$ & $g$ & $e$ & $\alpha$ \\
\midrule
Linear & 0.7 & -- & -- & -- & -- & -- & -- \\
SLERP & -- & 0.6 & -- & -- & -- & -- & -- \\
TIES & -- & -- & 0.2 & 0.7 & -- & -- & -- \\
DARE-TIES & -- & -- & 0.7 & 1.0 & -- & -- & -- \\
RegMean & 0.9 & -- & -- & -- & -- & -- & -- \\
LEWIS-TIES & -- & -- & -- & -- & 0.2 & 0.5 & -- \\
LEWIS-DARE-TIES & -- & -- & -- & -- & 0.4 & 0.7 & -- \\
AMARIS & -- & -- & -- & -- & -- & -- & 0.2 \\
\bottomrule
\end{tabular}
\end{table}

\begin{table}[t]
\centering
\small
\caption{32B SEA hyperparameters. Tuned methods use matched five-configuration validation-loss selection; untuned methods use the default settings for Table~\ref{tab:sea_untuned}. AMARIS fixes $\beta=1$.}
\label{tab:hparams_sea_appendix}
\begin{tabular}{lccccccc}
\toprule
Method & $w$ & $t$ & $d$ & $s$ & $g$ & $e$ & $\alpha$ \\
\midrule
\multicolumn{8}{l}{\textit{Validation-tuned methods}} \\
Linear & 0.1 & -- & -- & -- & -- & -- & -- \\
RegMean & 0.2 & -- & -- & -- & -- & -- & -- \\
AMARIS & -- & -- & -- & -- & -- & -- & 0.001 \\
\midrule
\multicolumn{8}{l}{\textit{Untuned methods}} \\
TIES & -- & -- & 0.2 & 1.0 & -- & -- & -- \\
DARE-TIES & -- & -- & 0.2 & 1.0 & -- & -- & -- \\
LEWIS-TIES & -- & -- & -- & -- & 0.3 & 0.8 & -- \\
LEWIS-DARE-TIES & -- & -- & -- & -- & 0.3 & 0.8 & -- \\
SLERP & -- & 0.5 & -- & -- & -- & -- & -- \\
\bottomrule
\end{tabular}
\end{table}


\section{SEAHELM Evaluation Implementation}
\label{app:seahelm_implementation}

\paragraph{Version and evaluation modes.}
We use SEAHELM v1 for the SEA evaluations reported in this paper. We manually extend the evaluation implementation to support non-thinking evaluation modes; all reported SEAHELM evaluations use non-thinking mode for every checkpoint.

\paragraph{Multi-turn judge.}
SEAHELM v1 includes a multi-turn task whose evaluation requires an OpenAI API-based judge. For this task, we use \texttt{gpt-oss-120b} \citep{gptoss} as the judge in our evaluation implementation.

\clearpage
\section*{NeurIPS Paper Checklist}

\begin{enumerate}

\item {\bf Claims}
    \item[] Question: Do the main claims made in the abstract and introduction accurately reflect the paper's contributions and scope?
    \item[] Answer: \answerYes{}
    \item[] Justification: The abstract and introduction report claim language calibrated to the measured Pareto operating points, and the quantitative statements are tied to reported experimental settings and tables (Abstract, Sections~\ref{sec:experiments} and \ref{sec:analysis_summary}).

\item {\bf Limitations}
    \item[] Question: Does the paper discuss the limitations of the work performed by the authors?
    \item[] Answer: \answerYes{}
    \item[] Justification: Section~\ref{sec:limitations} discusses dependence on calibration data, sensitivity to design choices, and the scope of pairwise merging with a shared architecture and tokenizer, including limitations for heterogeneous checkpoints, multi-specialist composition, and iterative merging.

\item {\bf Theory assumptions and proofs}
    \item[] Question: For each theoretical result, does the paper provide the full set of assumptions and a complete (and correct) proof?
    \item[] Answer: \answerNA{}
    \item[] Justification: The paper provides an objective derivation and algorithmic construction rather than theorem-proof style theoretical guarantees; derivation details are included in Appendix~\ref{app:setup_derivation}.

    \item {\bf Experimental result reproducibility}
    \item[] Question: Does the paper fully disclose all the information needed to reproduce the main experimental results of the paper to the extent that it affects the main claims and/or conclusions of the paper (regardless of whether the code and data are provided or not)?
    \item[] Answer: \answerYes{}
    \item[] Justification: The manuscript describes the merge algorithm, calibration datasets, evaluation protocol, selected hyperparameters, and bootstrap procedure (Sections~\ref{sec:method} and~\ref{sec:experiments}, and Appendices~\ref{app:selected_hparams} and~\ref{app:bootstrap}). Code is publicly available at \url{https://github.com/adriankhy93/AMARIS-neurips}.

\item {\bf Open access to data and code}
    \item[] Question: Does the paper provide open access to the data and code, with sufficient instructions to faithfully reproduce the main experimental results, as described in supplemental material?
    \item[] Answer: \answerNo{}
    \item[] Justification: Code is publicly available at \url{https://github.com/adriankhy93/AMARIS-neurips}. The manuscript identifies the source datasets and describes calibration-data construction, but we do not claim that all generated calibration data and artifacts needed to reproduce every reported experiment are publicly available.

\item {\bf Experimental setting/details}
    \item[] Question: Does the paper specify all the training and test details (e.g., data splits, hyperparameters, how they were chosen, type of optimizer) necessary to understand the results?
    \item[] Answer: \answerYes{}
    \item[] Justification: The paper specifies evaluation settings, calibration data construction, split/tuning protocols, and selected hyperparameters (Section~\ref{sec:experiments} and Appendix~\ref{app:selected_hparams}). The main 32B SEA comparison uses complete-task SEAHELM evaluation and matched five-configuration validation-loss tuning for AMARIS, RegMean, and Linear. Additional untuned SEA baselines are reported separately in Appendix~\ref{app:sea_untuned}.

\item {\bf Experiment statistical significance}
    \item[] Question: Does the paper report error bars suitably and correctly defined or other appropriate information about the statistical significance of the experiments?
    \item[] Answer: \answerYes{}
    \item[] Justification: We report finalized percentile 95\% confidence intervals from 10,000 paired evaluation-set bootstrap resamples for the 8B and 32B comparisons (Section~\ref{sec:statistical_uncertainty} and Appendix~\ref{app:bootstrap}). The appendix specifies the resampling units and metric recomputation. These intervals quantify evaluation-set sampling uncertainty for fixed checkpoints, not seed, decoding, or calibration variability.

\item {\bf Experiments compute resources}
    \item[] Question: For each experiment, does the paper provide sufficient information on the computer resources (type of compute workers, memory, time of execution) needed to reproduce the experiments?
    \item[] Answer: \answerNo{}
    \item[] Justification: Table~\ref{tab:runtime} reports measured stage-wise and total merge-pipeline runtime, peak GPU memory, and hardware at 8B and 32B scales; basis construction and greedy selection are timed jointly. Full accounting for calibration-completion generation, evaluation, hyperparameter sweeps, and total research compute is not provided.
    
\item {\bf Code of ethics}
    \item[] Question: Does the research conducted in the paper conform, in every respect, with the NeurIPS Code of Ethics \url{https://neurips.cc/public/EthicsGuidelines}?
    \item[] Answer: \answerYes{}
    \item[] Justification: The work uses established public assets, reports limitations and scope boundaries, and does not involve human-subject intervention experiments.

\item {\bf Broader impacts}
    \item[] Question: Does the paper discuss both potential positive societal impacts and negative societal impacts of the work performed?
    \item[] Answer: \answerNo{}
    \item[] Justification: The current draft focuses on method and benchmark behavior; a dedicated broader-impact discussion is not yet included.
    
\item {\bf Safeguards}
    \item[] Question: Does the paper describe safeguards that have been put in place for responsible release of data or models that have a high risk for misuse (e.g., pre-trained language models, image generators, or scraped datasets)?
    \item[] Answer: \answerNA{}
    \item[] Justification: This submission does not introduce a new open release of high-risk generative model weights or scraped datasets requiring additional release safeguards.

\item {\bf Licenses for existing assets}
    \item[] Question: Are the creators or original owners of assets (e.g., code, data, models), used in the paper, properly credited and are the license and terms of use explicitly mentioned and properly respected?
    \item[] Answer: \answerNo{}
    \item[] Justification: We cite the primary assets and sources, but explicit per-asset license and terms-of-use statements are not yet comprehensively enumerated in the manuscript.

\item {\bf New assets}
    \item[] Question: Are new assets introduced in the paper well documented and is the documentation provided alongside the assets?
    \item[] Answer: \answerNA{}
    \item[] Justification: The submission does not claim a new benchmark or dataset release as a primary contribution.

\item {\bf Crowdsourcing and research with human subjects}
    \item[] Question: For crowdsourcing experiments and research with human subjects, does the paper include the full text of instructions given to participants and screenshots, if applicable, as well as details about compensation (if any)? 
    \item[] Answer: \answerNA{}
    \item[] Justification: The experiments do not involve crowdsourcing or human-subject data collection.

\item {\bf Institutional review board (IRB) approvals or equivalent for research with human subjects}
    \item[] Question: Does the paper describe potential risks incurred by study participants, whether such risks were disclosed to the subjects, and whether Institutional Review Board (IRB) approvals (or an equivalent approval/review based on the requirements of your country or institution) were obtained?
    \item[] Answer: \answerNA{}
    \item[] Justification: No human-subject experiments were conducted.

\item {\bf Declaration of LLM usage}
    \item[] Question: Does the paper describe the usage of LLMs if it is an important, original, or non-standard component of the core methods in this research? Note that if the LLM is used only for writing, editing, or formatting purposes and does \emph{not} impact the core methodology, scientific rigor, or originality of the research, declaration is not required.
    \item[] Answer: \answerYes{}
    \item[] Justification: Section~\ref{sec:experiments} and Appendix~\ref{app:reproducibility} describe the use of the source models to generate calibration responses. Appendix~\ref{app:seahelm_implementation} documents the substitution of \texttt{gpt-oss-120b} as the SEAHELM multi-turn judge. SEA-Instruct calibration uses existing question--answer pairs without additional response generation.

\end{enumerate}

\end{document}